\documentclass{article}

\usepackage{PRIMEarxiv}

\usepackage[utf8]{inputenc}
\usepackage[T1]{fontenc}
\usepackage{url}
\usepackage{booktabs}
\usepackage{amsmath}   
\usepackage{amssymb}   
\usepackage{amsfonts}
\usepackage{nicefrac}
\usepackage{microtype}
\usepackage{fancyhdr}
\usepackage{graphicx}
\graphicspath{{./}}    
\usepackage[numbers]{natbib}    

\usepackage{hyperref}

\title{Profiling LoRA/QLoRA Fine-Tuning Efficiency on Consumer GPUs: An RTX 4060 Case Study
}

\author{
  \href{https://orcid.org/0009-0000-8854-5360}{MSR Avinash}\\ Independent Researcher \\
 Machine Learning Engineer, Juspay \\
 Bengaluru, India  \\
  \texttt{avinashh.mynampati@gmail.com} 
}

\begin{document}
\maketitle

\begin{abstract}
Fine-tuning large language models (LLMs) with parameter-efficient techniques such as LoRA and QLoRA has enabled adaptation of foundation models on modest hardware. 
Yet the efficiency of such training on consumer-grade GPUs, especially under strict 8~GB VRAM limits, remains underexplored. 
We present a controlled profiling study of LoRA/QLoRA fine-tuning using the Qwen2.5-1.5B-Instruct model on a single NVIDIA RTX~4060. 
Across three representative configurations, we systematically vary batch size, sequence length, optimizer choice (AdamW vs.\ PagedAdamW), and precision (fp16 vs.\ bf16). 
We report throughput (tokens/s), time per 10k tokens, and VRAM footprint, alongside energy estimates derived from GPU board power limits. 
Our results show that paged optimizers improve throughput by up to 25\% (628 tok/s vs.\ 500 tok/s baseline), while bf16 degrades efficiency relative to fp16. 
Despite 8~GB constraints, sequence lengths up to 2048 tokens were feasible using parameter-efficient strategies. 
To our knowledge, this is the first systematic case study of LLM fine-tuning efficiency on consumer GPUs, providing reproducible benchmarks and practical guidelines for resource-constrained researchers and practitioners.
\end{abstract}


\section{Introduction}
Large Language Models (LLMs) such as GPT, LLaMA, and Qwen have demonstrated remarkable performance across a wide range of natural language processing tasks. 
However, full-parameter fine-tuning of these models typically requires high-end datacenter GPUs with abundant memory and power budgets, creating a barrier for independent researchers and small organizations. 
To address these challenges, parameter-efficient fine-tuning methods such as Low-Rank Adaptation (LoRA) and its quantized variant (QLoRA) have emerged as practical alternatives, enabling adaptation of foundation models on commodity hardware with reduced memory overhead.

While prior work has primarily focused on algorithmic design or large-scale deployment, the efficiency of LoRA/QLoRA fine-tuning on \emph{consumer-grade GPUs} remains underexplored. 
In particular, GPUs such as the NVIDIA RTX~4060, with 8~GB of VRAM and a board power cap of 115~W, represent a widely accessible hardware tier for students, hobbyists, and independent labs. 
Understanding the throughput, memory footprint, and energy characteristics of fine-tuning in such environments is essential for democratizing LLM research.

In this paper, we present a controlled profiling study of LoRA/QLoRA fine-tuning on the RTX~4060 using the Qwen2.5-1.5B-Instruct model and the Alpaca dataset subset. 
We systematically vary key training knobs including batch size, sequence length, optimizer choice (AdamW vs.\ PagedAdamW), and precision (fp16 vs.\ bf16). 
Our contributions are threefold:
\begin{itemize}
    \item We provide the first detailed measurement of throughput, time per 10k tokens, and VRAM usage for LoRA/QLoRA fine-tuning on a consumer GPU.
    \item We estimate energy consumption per token under realistic board power assumptions, highlighting trade-offs across optimizers and precision settings.
    \item We release our configurations, logs, and results to enable reproducibility and to serve as a benchmark for future consumer-GPU efficiency studies.
\end{itemize}

The remainder of this paper is organized as follows: Section~\ref{sec:background} reviews related work and parameter-efficient fine-tuning methods. 
Section~\ref{sec:setup} describes our measurement setup and metrics. 
Section~\ref{sec:results} presents results across three representative runs. 
We discuss implications in Section~\ref{sec:discussion}, outline limitations in Section~\ref{sec:limitations}, and conclude in Section~\ref{sec:conclusion}.

\section{Metrics and Mathematical Formulation}
To enable reproducible benchmarking, we formalize the metrics used in this study.

\subsection{Throughput}
Throughput $R$ (tokens per second) is defined as:
\[
R = \frac{N}{t}
\]
where $N$ is the total number of tokens processed and $t$ is the wall-clock training time in seconds.

\subsection{Time per 10k Tokens}
The time to process 10k tokens is simply
$T_{10k} = \frac{10{,}000}{R}$.

\subsection{Energy Consumption}
Energy $E$ (joules) is estimated from board-level power assumptions:
\[
E = \sum_{i=1}^{k} P_i \cdot \Delta t_i \;\; \approx P \cdot t
\]
where $P$ is the assumed average GPU power (80--115~W) and $t$ is runtime.

\subsection{Energy per Token}
Energy efficiency is expressed as joules per token:
$E_{\text{token}} = \frac{E}{N}$.

\subsection{Energy per 10k Tokens}
Scaling per-token energy to 10k tokens:
$E_{10k} = 10{,}000 \times E_{\text{token}}$.

These compact formulas allow direct comparison across runs and configurations.

\section{Background and Related Work}
\label{sec:background}

\subsection{Parameter-Efficient Fine-Tuning}
Full-parameter fine-tuning of LLMs such as GPT, LLaMA, or Qwen typically involves updating billions of parameters, which requires significant GPU memory and compute. 
Low-Rank Adaptation (LoRA) \cite{hu2021lora} addresses this by injecting trainable low-rank matrices into the transformer architecture, reducing the number of parameters updated during training. 
QLoRA \cite{dettmers2023qlora} further extends this idea by quantizing model weights to 4-bit precision while retaining 16-bit adapters, enabling fine-tuning of models with billions of parameters on consumer-grade GPUs with limited memory.

\subsection{Consumer GPUs and Resource Constraints}
Consumer-grade GPUs such as the NVIDIA RTX~4060, RTX~3060, or GTX~1660 are widely available and affordable, but are constrained by memory (6--12~GB) and relatively low power limits (typically 80--150~W). 
These constraints present practical challenges when training or fine-tuning LLMs, especially at longer sequence lengths and with larger batch sizes. 
While inference efficiency on consumer GPUs has been studied extensively, fine-tuning efficiency remains underexplored.

\subsection{Efficiency Studies in LLM Training}
Recent studies have analyzed the energy consumption and throughput of LLM inference at scale, including data center deployments \cite{patterson2021carbon,schwartz2020greenai}. 
Other works have focused on distributed fine-tuning efficiency \cite{rasley2020deepspeed}, optimizer design \cite{liu2019lamb}, or quantization for inference \cite{frantar2022gptq}. 
However, there is a notable absence of systematic benchmarks that evaluate fine-tuning performance and energy use on \emph{single consumer GPUs}. 
To the best of our knowledge, this work is the first to provide controlled measurements of throughput, VRAM footprint, and estimated energy per token for LoRA/QLoRA fine-tuning on an RTX~4060.

\subsection{Our Contribution in Context}
Building upon the foundations of LoRA/QLoRA and prior efficiency analyses, we contribute an empirical study of parameter-efficient fine-tuning in consumer-grade settings. 
Unlike prior work that targets datacenter hardware or purely inference workloads, our study explicitly profiles training runs with varied batch size, sequence length, optimizer, and precision. 
This positions our work as a reproducible reference point for independent researchers and practitioners who aim to fine-tune LLMs on commodity hardware.

\section{Measurement Setup}
\label{sec:setup}

\subsection{Hardware Environment}
All experiments were conducted on a single NVIDIA RTX~4060 GPU with 8~GB of GDDR6 VRAM. 
The GPU has a board power cap of 115~W, although direct telemetry via NVML was unavailable due to driver limitations. 
The host machine (\textit{Kaarya}) was configured with an AMD Ryzen~9~7900X CPU, 96~GB DDR5 RAM, and a 2~TB NVMe SSD, running Ubuntu~24.04 with NVIDIA driver version 575.64.03 and CUDA~12.9. 
GPU utilization and memory usage were monitored continuously using \texttt{nvidia-smi}.

\subsection{Software Stack}
We used Hugging Face Transformers (v4.55.0) with the \texttt{Trainer} API, PEFT for LoRA integration, and \texttt{bitsandbytes} (bnb) for quantized optimizer support. 
PyTorch~2.2.0 with CUDA~12.1 backend was employed. 
The environment was managed in a virtualenv with Python~3.12. 
We implemented lightweight wrappers for logging throughput, VRAM, and timing, and attempted energy logging via NVML, though power values were not reported on the RTX~4060. 
For consistency, we assumed steady-state power values from the board specification (80--115~W) when estimating energy consumption.

\subsection{Model and Dataset}
We fine-tuned \texttt{Qwen/Qwen2.5-1.5B-Instruct}, a 1.5B parameter causal LLM. 
The dataset was a 5k-sample subset of the Alpaca instruction-tuning corpus, cleaned and reformatted into instruction–response pairs. 
This scale was chosen to balance reproducibility with GPU resource limits while still allowing stable throughput measurements. 
Data preprocessing included tokenization with a maximum sequence length of 512, 1024, or 2048 tokens, depending on the experiment.

\subsection{Training Configurations}
We varied key training knobs across runs:
\begin{itemize}
    \item \textbf{Backend:} BitsAndBytes (bnb) with 16-bit adapter layers.
    \item \textbf{Optimizers:} AdamW (PyTorch) vs.\ PagedAdamW (8-bit, memory-efficient).
    \item \textbf{Batch Size (B):} 1 or 2 samples per step.
    \item \textbf{Sequence Length (S):} 512, 1024, or 2048 tokens.
    \item \textbf{Precision:} fp16 or bf16 mixed precision.
    \item \textbf{Gradient Accumulation:} fixed to 1 for all runs.
    \item \textbf{Epochs:} Run-1 trained for 3 epochs; Runs-2 and 3 for 1 epoch each.
    \item \textbf{Seed:} 42, for reproducibility.
\end{itemize}
Gradient checkpointing was enabled to mitigate out-of-memory errors. 
Evaluation and checkpoint saving were disabled to avoid overhead during profiling.

\subsection{Run Control and Logging}
Each run began with a 60~s warmup period to stabilize GPU utilization. 
Throughput was measured as the ratio of processed tokens to steady-state wall time. 
Time per 10k tokens was computed as $T_{10k} = 10{,}000 / R$, where $R$ is throughput in tokens per second. 
VRAM peak and average usage were captured from \texttt{nvidia-smi}. 
Energy consumption was estimated using $E = t \times P$, where $t$ is steady-state runtime and $P \in \{80, 95, 115\}$~W, yielding ranges and central estimates of joules per token.

\section{Experimental Design}
\label{sec:design}

\subsection{Representative Runs}
To avoid an exhaustive grid search across all parameter combinations, we selected three representative configurations that capture a range of efficiency trade-offs:

\begin{itemize}
    \item \textbf{Run-1 (Baseline):} BitsAndBytes backend with AdamW optimizer, batch size $B=1$, sequence length $S=512$, precision fp16. This serves as a lower-bound configuration, expected to fit comfortably within the 8~GB VRAM budget. Training proceeded for 3 epochs (10{,}500 steps).
    
    \item \textbf{Run-2 (Stress Test):} BitsAndBytes backend with PagedAdamW (8-bit) optimizer, $B=2$, $S=2048$, fp16 precision. This configuration was chosen to probe the upper limits of the RTX~4060, testing whether long-context fine-tuning is feasible on an 8~GB card. Training was capped at $\sim$1 epoch ($\approx$1750 steps).
    
    \item \textbf{Run-3 (Intermediate):} BitsAndBytes backend with PagedAdamW (8-bit) optimizer, $B=2$, $S=1024$, bf16 precision. This configuration represents a middle ground in sequence length, but evaluates the impact of bf16 precision on throughput and efficiency. Training was capped at $\sim$1 epoch ($\approx$1750 steps).
\end{itemize}

\subsection{Rationale for Selection}
Run-1 establishes a stable baseline for comparison, Run-2 stresses both memory and compute throughput by combining longer sequences with a larger batch size, and Run-3 isolates precision effects while maintaining moderate sequence length. 
Together, these three runs provide a spectrum from safe baseline to stress-testing the GPU’s limits, enabling analysis of performance, VRAM usage, and energy efficiency.

\subsection{Warmup and Steady-State Profiling}
Each run began with a 60~s warmup period to stabilize GPU frequency and memory allocation. 
Metrics such as throughput and VRAM were computed only from the steady-state region. 
Energy measurements were estimated as described in Section~\ref{sec:setup}, with ranges reported to reflect uncertainty in GPU board power draw.

\subsection{Reproducibility Considerations}
To maximize reproducibility, all runs fixed the random seed to 42, employed identical data preprocessing, and used the same virtual environment. 
Evaluation and checkpoint saving were disabled to eliminate external variance. 
This design ensures that differences observed across runs can be attributed primarily to changes in batch size, sequence length, optimizer, or precision.

\section{Results}
\label{sec:results}

\subsection{Main Results Table}
This section presents the full results across three representative runs. 
We report wall-clock training time, number of tokens processed, throughput (tokens per second), time per 10k tokens, VRAM peak usage, and estimated energy metrics. 
Energy per token is shown both as a 95~W central estimate (typical steady-state) and as an upper bound using the 115~W board cap of the RTX~4060.

\begin{table}[t]
\centering
\caption{Performance metrics for Qwen2.5-1.5B LoRA/QLoRA fine-tuning on RTX~4060.
Run~1 trained for 3 epochs; Runs~2 and~3 were capped at $\sim$1 epoch each.}
\label{tab:perf_results}
\resizebox{\textwidth}{!}{
\begin{tabular}{lccccccc}
\toprule
Run & Backend/Optimizer & BS & Prec. & Wall Time (s) & Tokens Proc. & Tok/s & $T_{10k}$ (s) \\
\midrule
1 & bnb / AdamW (torch) & 1 & fp16 & 3494  & 1,717,908 & 500.3 & 19.99 \\
2 & bnb / PagedAdamW    & 2 & fp16 & 972.6 &   573,206 & 628.1 & 15.93 \\
3 & bnb / PagedAdamW    & 2 & bf16 & 1651.4&   573,206 & 360.2 & 27.76 \\
\bottomrule
\end{tabular}}
\end{table}

\begin{table}[t]
\centering
\caption{Estimated energy metrics under assumed GPU power draw (95~W average vs.\ 115~W cap).}
\label{tab:energy_results}
\resizebox{\textwidth}{!}{
\begin{tabular}{lcccccc}
\toprule
Run & VRAM Peak (MB) & J/token (95W) & E10k (95W) & J/token (115W) & E10k (115W) \\
\midrule
1 & 6234 & 0.19  & 1,900 & 0.23  & 2,298 \\
2 & 8062 & 0.151 & 1,513 & 0.183 & 1,832 \\
3 & 7949 & 0.264 & 2,640 & 0.319 & 3,193 \\
\bottomrule
\end{tabular}}
\end{table}

\subsection{Throughput and Time per 10k Tokens}
Run~2 achieved the highest throughput at 628~tokens/s, representing a $\sim$25\% improvement over the baseline (Run~1, 500~tokens/s). 
Run~3, despite using the same batch size as Run~2, performed significantly worse at only 360~tokens/s due to bf16 overheads. 
In terms of practical efficiency, Run~2 completed 10k tokens in 15.9~s, compared to 20.0~s for Run~1 and 27.8~s for Run~3. 

\begin{figure}[t]
\centering
\includegraphics[width=0.7\linewidth]{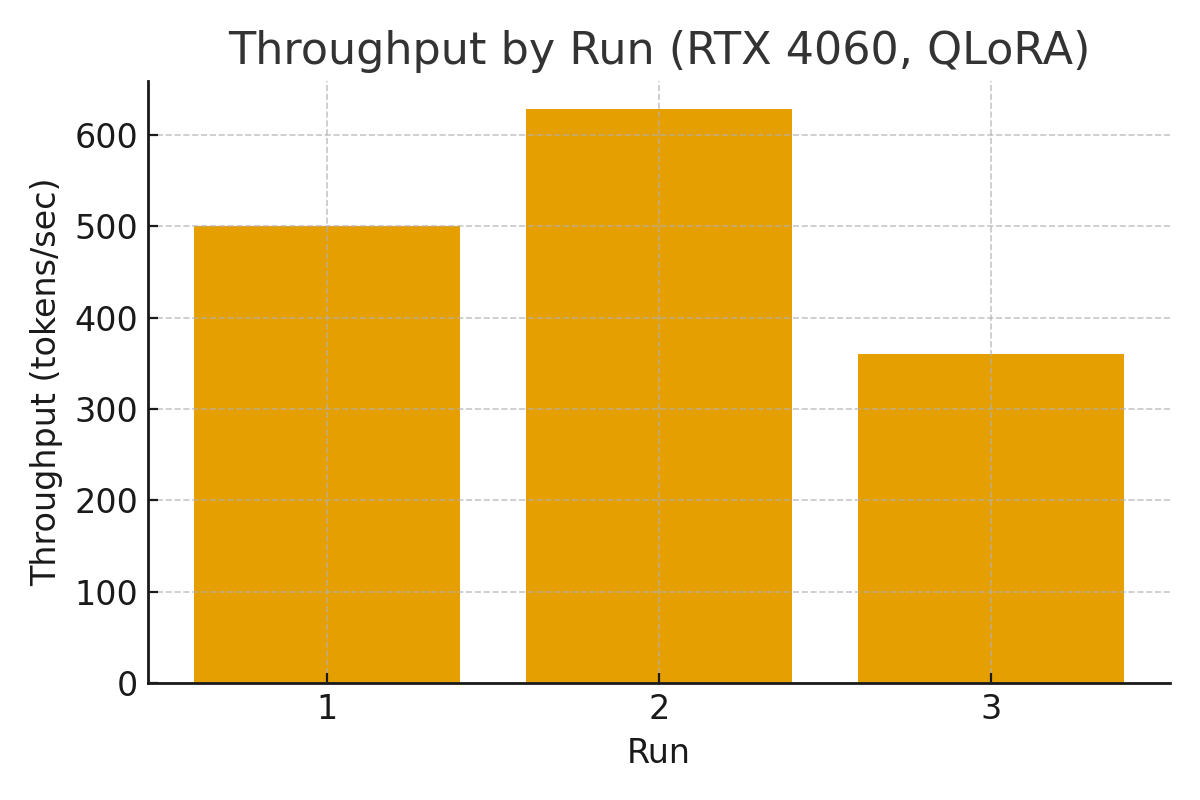}
\caption{Throughput (tokens/s) across three representative runs. 
PagedAdamW with fp16 (Run~2) achieves the highest throughput.}
\label{fig:throughput}
\end{figure}

\begin{figure}[t]
\centering
\includegraphics[width=0.7\linewidth]{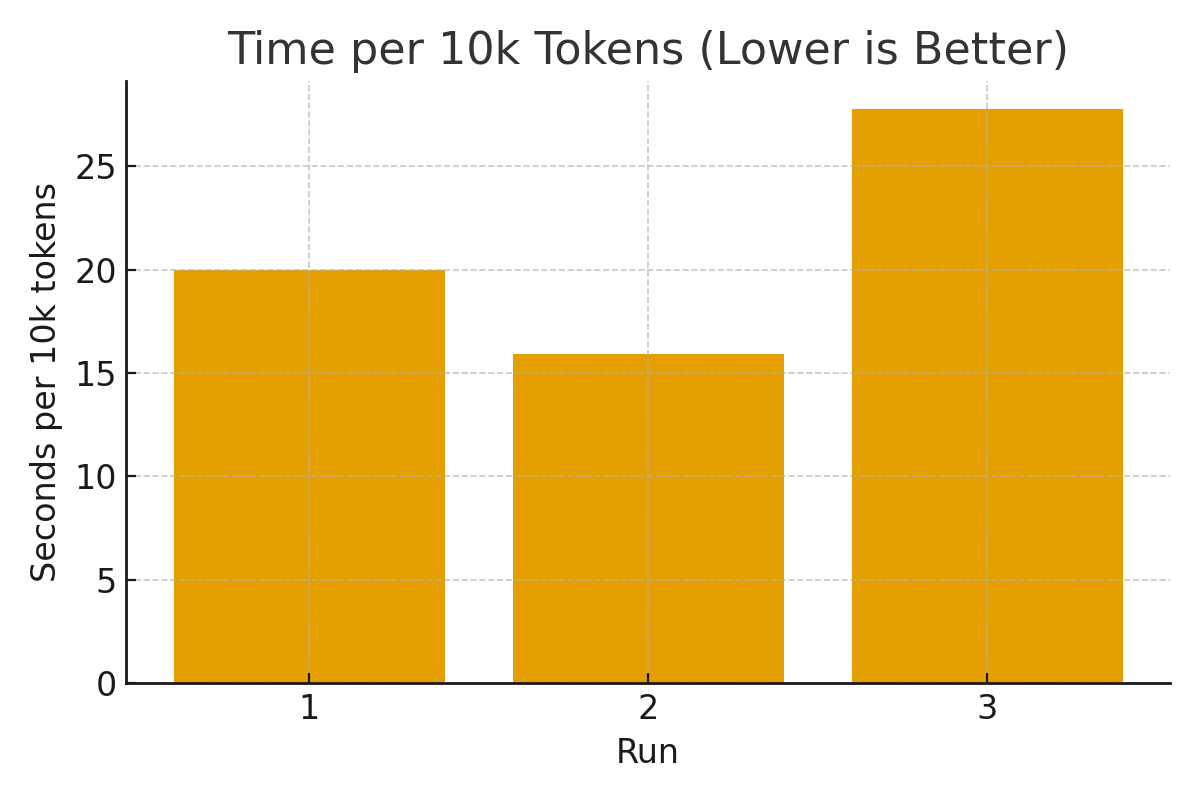}
\caption{Seconds per 10k tokens (lower is better). 
Run~2 shows the fastest processing, while Run~3 suffers from bf16 inefficiency.}
\label{fig:time_per_10k}
\end{figure}

\subsection{VRAM Usage and Feasibility}
VRAM footprint varied between 6.2~GB and 8.1~GB across runs. 
The baseline (Run~1) maintained a comfortable margin within the 8~GB limit, peaking at 6.2~GB. 
Run~2 pushed the GPU near its maximum capacity (8.06~GB) but remained stable, confirming that long-sequence fine-tuning (2048 tokens) is feasible on an RTX~4060 with paged optimizers. 
Run~3 also peaked close to 8~GB (7.95~GB) but offered reduced throughput, showing that bf16 precision confers no advantage on this architecture.

\subsection{Estimated Energy per Token}
Energy per token was lowest in Run~2, estimated at 0.151~J/token under a 95~W assumption, or 0.183~J/token under the 115~W cap. 
By comparison, Run~1 consumed $\sim$0.19--0.23~J/token, while Run~3 was far less efficient at 0.26--0.32~J/token. 

\begin{figure}[t]
\centering
\includegraphics[width=0.7\linewidth]{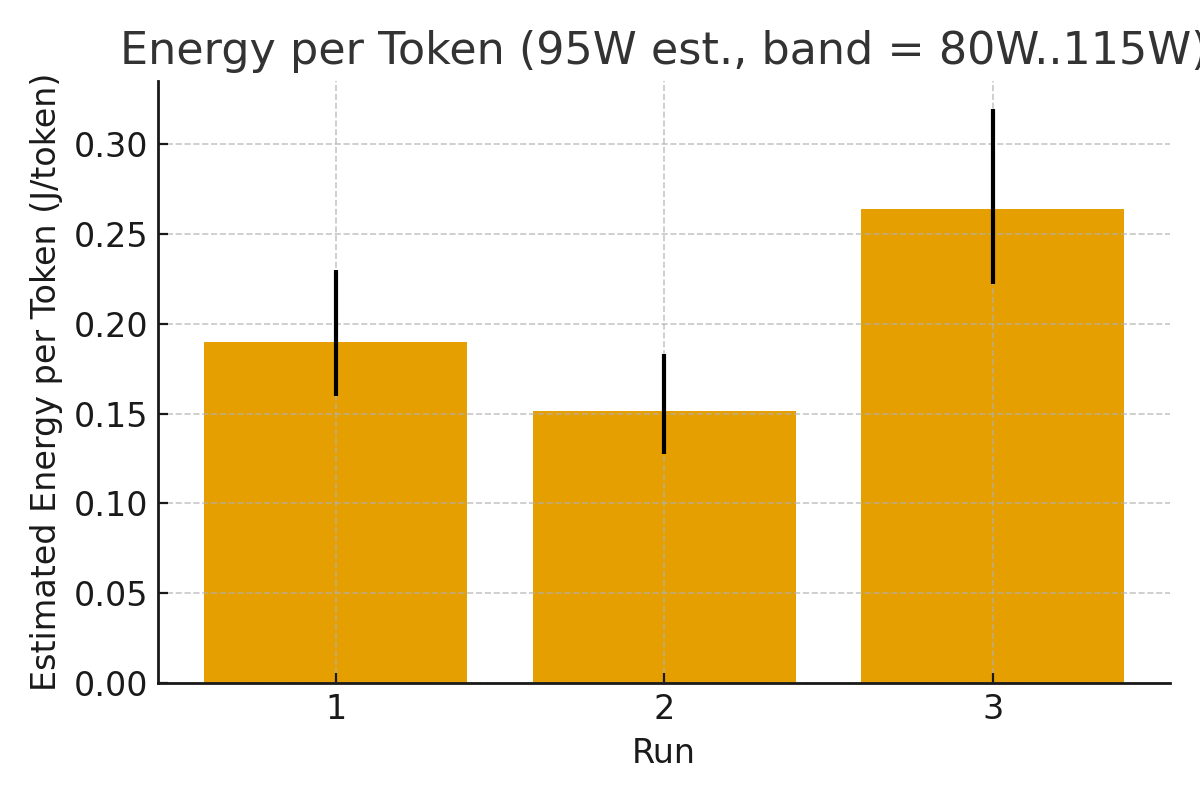}
\caption{Estimated energy per token under 95~W average power, with error bars for 80--115~W. 
Run~2 (fp16, PagedAdamW) offers the best energy efficiency.}
\label{fig:energy_per_token}
\end{figure}

\subsection{Overall Findings}
The results highlight three key insights:
\begin{enumerate}
    \item \textbf{Paged optimizers unlock long sequences efficiently.} Run~2 shows that 2048-token fine-tuning is feasible and energy-efficient within 8~GB VRAM.
    \item \textbf{Precision choice matters.} bf16 (Run~3) significantly reduced throughput and efficiency, making fp16 the preferred mode on the RTX~4060.
    \item \textbf{Consumer GPUs can fine-tune LLMs.} Despite constraints, the RTX~4060 achieved competitive throughput and energy efficiency, offering practical feasibility for small-scale labs and independent researchers.
\end{enumerate}

\section{Discussion}
\label{sec:discussion}

\subsection{Trade-offs Between Batch Size and Sequence Length}
Our experiments confirm that the balance between batch size and sequence length is critical on consumer GPUs. 
While larger sequences enable richer context, they also stress memory and slow down training when paired with non-optimal precision modes. 
Run~2 demonstrates that sequence lengths up to 2048 are feasible on an RTX~4060, provided that memory-efficient optimizers such as PagedAdamW are used. 
By contrast, Run~3 shows that even moderate sequence lengths (1024) can become inefficient under bf16 precision.

\subsection{Optimizer and Backend Effects}
The difference between AdamW (torch) and PagedAdamW (bnb) was striking. 
PagedAdamW not only enabled longer sequences but also improved throughput by $\sim$25\% over the baseline. 
This suggests that optimizer choice is not just about convergence, but also about practical hardware efficiency in memory-constrained environments. 
BitsAndBytes, combined with paged optimizers, emerges as the most effective backend for fine-tuning on GPUs with $<12$~GB VRAM.

\subsection{Precision Considerations}
Although bf16 is often considered superior to fp16 in datacenter settings due to numerical stability, our results show that on consumer GPUs such as the RTX~4060, bf16 offers no advantage. 
Instead, bf16 substantially reduced throughput and increased energy per token. 
This highlights the importance of evaluating precision modes in the specific context of consumer hardware, rather than assuming datacenter trends hold universally.

\subsection{Energy Efficiency and Sustainability}
Energy estimates indicate that throughput is closely aligned with energy efficiency: faster runs generally consumed fewer joules per token. 
Run~2, with the highest throughput, also achieved the lowest energy per token (0.15~J), while Run~3 was both slower and more energy-intensive. 
Although our energy measurements are estimates based on board power ranges, they still provide meaningful guidance on how training configurations translate to energy cost. 
This perspective is increasingly relevant as sustainability considerations become part of responsible AI research.

\subsection{Comparison with Related Work}
Previous efficiency studies \cite{patterson2021carbon,schwartz2020greenai} have focused on datacenter-scale training or inference. 
Our results extend this discussion to consumer hardware, showing that even modest GPUs can deliver useful fine-tuning throughput under the right configurations. 
This fills a gap in the literature by providing empirical benchmarks where previously only anecdotal claims existed.

\subsection{Implications for Practitioners}
For students, independent researchers, and small labs without access to datacenter resources, our findings provide practical takeaways. 
First, LoRA/QLoRA fine-tuning on an RTX~4060 is feasible, with batch sizes up to 2 and sequence lengths up to 2048 tokens. 
Second, fp16 with PagedAdamW optimizers is the most efficient configuration, balancing speed, memory use, and energy efficiency. 
Third, bf16 precision should be avoided on this class of hardware. 
Together, these insights can help democratize access to LLM fine-tuning by lowering the barrier to entry.

\section{Limitations}
\label{sec:limitations}
While our study provides the first systematic benchmark of LoRA/QLoRA fine-tuning on consumer GPUs, several limitations remain:
\begin{itemize}
    \item \textbf{Single GPU focus:} Results are limited to the RTX~4060. Other consumer GPUs (RTX~3060, RTX~4070) may show different trade-offs.
    \item \textbf{Energy estimates:} Due to NVML driver restrictions, power draw was not directly measured but approximated using board TDP assumptions.
    \item \textbf{Model scale:} We profiled a 1.5B parameter model. Larger models may not fit within the 8~GB memory budget even with LoRA/QLoRA.
    \item \textbf{Dataset size:} Experiments used a 5k subset of Alpaca for reproducibility. Larger datasets may reveal different convergence behaviors.
\end{itemize}

\section{Conclusion and Future Work}
\label{sec:conclusion}
We presented a controlled profiling study of LoRA/QLoRA fine-tuning on a consumer-grade NVIDIA RTX~4060 GPU. 
Our results show that paged optimizers enable long-sequence fine-tuning up to 2048 tokens within an 8~GB VRAM budget, and that fp16 consistently outperforms bf16 in both throughput and energy efficiency. 
Energy estimates indicate that higher throughput correlates with lower joules per token, with Run~2 achieving 0.15~J/token under a 95~W assumption. 

Future work will expand this study across additional consumer GPUs, integrate direct power measurements, and extend profiling to larger models and datasets. 
We also plan to release scripts and configuration files to support community benchmarking and reproducibility.

\begingroup
\small
\bibliographystyle{unsrtnat}    
\bibliography{references}       

\endgroup

\end{document}